\documentclass[runningheads]{llncs}
\usepackage{bm}
\usepackage[T1]{fontenc}
\usepackage{graphicx,verbatim}
\usepackage{amssymb}
\usepackage[mathscr]{eucal}
\usepackage{mathrsfs}
\usepackage{bm}
\usepackage{multirow}
\usepackage{booktabs}
\usepackage{makecell}
\usepackage{bbding}
\usepackage{graphicx}
\usepackage[table,xcdraw]{xcolor}
\usepackage{graphicx}
\usepackage{float}
\usepackage[colorlinks,hyperindex,breaklinks]{hyperref}
\usepackage{hyperref}
\usepackage{graphicx} 
\usepackage{float} 
\usepackage{subfigure} 

\usepackage{marvosym}

\newcommand{\captab}{-2.mm}

\newcommand{\tabbottom}{-0.mm}

\begin{document}
\title{CoPA: Hierarchical Concept Prompting and Aggregating Network for Explainable Diagnosis} 

\author{Yiheng Dong\inst{1}$^{*}$ 
\and
Yi Lin\inst{2}$^{*}$ 
\and
Xin Yang\inst{1}\textsuperscript{{(}\Letter{)}}} 

\institute{
School of Electronic Information and Communications, Huazhong University of Science and Technology, Wuhan, China \\
\email{xinyang2014@hust.edu.cn}
\and
Department of Computer Science and Engineering, The Hong Kong University of Science and Technology, Hong Kong, China
}
\authorrunning{Y. Dong et al.}
%
\titlerunning{CoPA: Concept Prompting and Aggregating Network.}
\renewcommand{\thefootnote}{\fnsymbol{footnote}}
\footnotetext[1]{These authors contributed equally to this work.}
    
\maketitle              
\begin{abstract}
The transparency of deep learning models is essential for clinical diagnostics. Concept Bottleneck Model provides clear decision-making processes for diagnosis by transforming the latent space of black-box models into human-understandable concepts. However, concept-based methods still face challenges in concept capture capabilities. These methods often rely on encode features solely from the final layer, neglecting shallow and multiscale features, and lack effective guidance in concept encoding, hindering fine-grained concept extraction. To address these issues, we introduce Concept Prompting and Aggregating (CoPA), a novel framework designed to capture multilayer concepts under prompt guidance. This framework utilizes the Concept-aware Embedding Generator (CEG) to extract concept representations from each layer of the visual encoder. Simultaneously, these representations serve as prompts for Concept Prompt Tuning (CPT), steering the model towards amplifying critical concept-related visual cues. Visual representations from each layer are aggregated to align with textual concept representations. With the proposed method, valuable concept-wise information in the images is captured and utilized effectively, thus improving the performance of concept and disease prediction. Extensive experimental results demonstrate that CoPA outperforms state-of-the-art methods on three public datasets. Code is available at \href{https://github.com/yihengd/CoPA}{https://github.com/yihengd/CoPA}.
\end{abstract}
\section{Introduction}
Rapid advancements in deep learning have led to remarkable progress in medical image analysis~\cite{litjens2017survey,azizi2021big,shamshad2023transformers}, yet ensuring model interpretability remains a critical challenge in clinical practice~\cite{XAI}. Traditional post-hoc interpretability methods, such as CAM~\cite{CAM} and Grad-CAM~\cite{GradCAM}, which provide visualizations of model decisions, often fail to meet the high precision and reliability demands of medical scenarios~\cite{laugel2019dangers,rudin2019stop}. To address this limitation, Concept Bottleneck Model (CBM)~\cite{CBM} has emerged as a promising solution. CBM maps input images to a predefined set of human-interpretable concepts, serving as a verifiable ``bridge'' between raw inputs and final predictions, making the decision-making process interpretable. These concepts encompass a spectrum of features, ranging from low-level attributes like color and shape to high-level semantic features, such as ulceration, providing deep insights into the model's decision-making process.

However, existing CBM implementations~\cite{PCBM,CEM,Explicd,MICA,CBE} often exhibit limited effectiveness in capturing concepts, as they typically rely on final-layer features of the image encoder (ResNet/ViT) for concept alignment. As noted by~\cite{raghu2021vision,xu2023fine}, although these deep representations often tend to capture high-level global semantics, they inevitably overlook critical low-level and local visual patterns. This representation deficiency leads to the inadequate encoding of concepts that require shallow or multiscale analysis (e.g., dots and globules), ultimately impacting concept alignment and disease diagnosis.

Recent integration of Vision-Language Models (VLMs) with CBMs has revealed new opportunities through their pre-trained cross-modal alignment capabilities~\cite{PCBM,Explicd,MICA,LaBo}. However, due to the scarcity of annotated medical datasets, VLMs face challenges in capturing fine-grained concept semantics, particularly for recognizing complex pathological patterns. Furthermore, the issue of model forgetting~\cite{forgetting2} warrants attention. After fine-tuning on downstream tasks, models may significantly forget previously encoded specialized medical knowledge.

To address these challenges, we propose \textbf{Co}ncept \textbf{P}rompting and \textbf{A}ggregating network (CoPA) aiming at enabling fine-grained and multiscale feature capture and differentiation for concepts. Specifically, Concept-aware Embedding Generator (CEG) is proposed to distill highly concentrated concept-aware feature representations, which are aggregated by a selector to form the final visual concept representation. Subsequently, we introduce Concept Prompt Tuning (CPT), where the outputs of CEG serve as inputs to the next transformer layer with the backbone frozen, guiding the concept prompt to progressively concentrate more on concept-related features. Contrastive learning is then utilized to align visual and textual concept representations. Finally, a gating network is employed to weigh and combine the aligned concept representations for disease prediction. 

The main contributions of our work are as follows: 1) We present Concept Prompting and Aggregating (CoPA), a novel explainable network adept at capturing multiscale and fine-grained visual concept representations. 2) We design Concept-aware Embedding Generator (CEG), which extracts highly concentrated concept-related visual representations from each layer of the visual encoder, facilitating multilayer feature aggregation. 3) We design Concept Prompt Tuning (CPT) technique to guide the model's focus on target visual concepts and mitigate the issue of knowledge forgetting. 4) Extensive experimental results demonstrate that our CoPA outperforms state-of-the-art methods, highlighting the efficacy of each component in our framework.
\section{Method}
\subsection{Overall Pipeline}
\begin{figure}[!t]
\centering\includegraphics[width=\textwidth]{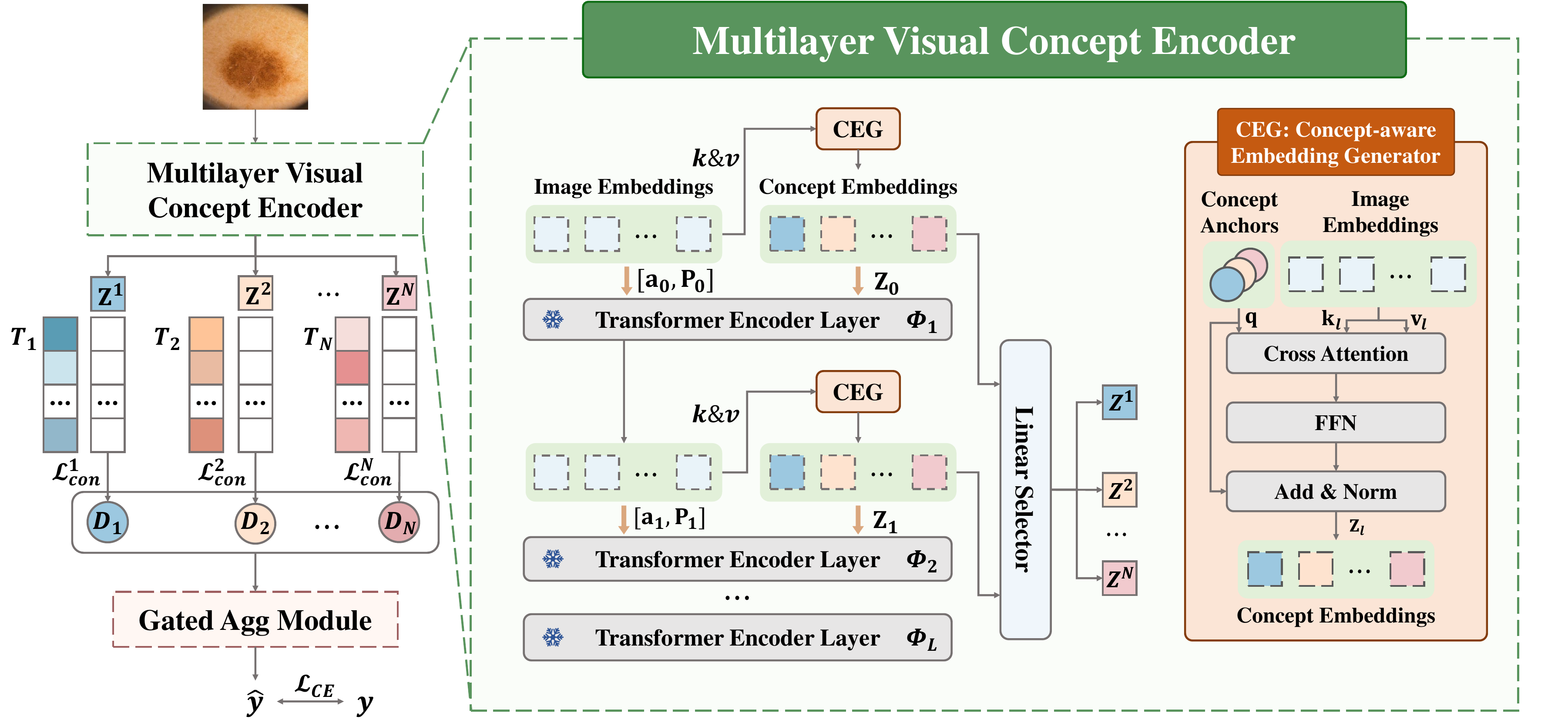}
\caption{The overall pipeline of CoPA, which consists of a multi-layer visual concept encoder, a concept alignment bottleneck layer, and a gated aggregation module.}
\label{fig:framework}
\end{figure}

The architecture of the proposed model is shown in Fig. \ref{fig:framework}. Given a data represented by triplets $\mathcal{D}=\{(x,c,y)\}$, where $x$ denotes the input image, $y$ represents the disease label and $c=\{c_1,c_2...c_N\}$ signifies a set of concept labels with the number of $N$. Moreover, each concept $c_i$ belongs to a candidate set $\mathcal{C}_i=\{c_i^1,c_i^2,...,c_i^{k_i}\}$ (e.g., for the concept ``Pigment Network'',~$\mathcal{C}_i=$ \{``atypical'', ``typical''\}), where 
${k_i}$ indicates the number of elements in $\mathcal{C}_i$ for the ${i^{th}}$ concept. 

 We start with passing the image $x$ through Multilayer Visual Concept Encoder to generate concept-aware visual embeddings. In parallel, text embeddings for each concept candidate set $\mathcal{C}_i$ are generated using a frozen text encoder. Afterwards, cross-modal alignment is achieved through contrastive learning between two modalities. Finally, the aligned representations undergo adaptive fusion and generate disease predictions.

\subsection{Multilayer Visual Concept Learning}
Visual concept representation extraction comprises two critical components operating at each visual encoder layer. Unlike some prior methods that predict all concepts using a single feature map, our framework leverages concept-aware visual embeddings from Concept-aware Embedding Generator (CEG) to effectively encode the relevant visual features associated with specific concepts. However, the model's capability to localize concept-aware visual information remains suboptimal. We attribute this to insufficient semantic guidance during encoding and catastrophic forgetting during full-parameter fine-tuning~\cite{forgetting2}. Therefore, Concept Prompt Tuning (CPT) is proposed to inject semantic guidance into feature extraction. Hierarchically aggregated layer-wise representations yield a unified visual concept embedding for cross-modal alignment.

 \textbf{Concept-aware Embedding Generator (CEG)}. Our CEG takes two sides of input: a set of learnable concept anchors and a token-wise feature map from the visual encoder layer. Each anchor ${\mathbf{q}_i}\in\mathbb{R}^{d}, 1\leq{\mathit{i}}\leq N$, where $N$ denotes the length of concept set $c$, is associated with a specific concept and serves as a query vector to retrieve essential information from visual features. For a given input from the $l^{th}$ encoder layer, we consider image embedding as both key and value vectors $\mathbf{k}_l,\mathbf{v}_l\in\mathbb{R}^{m\times d}$, where $m$ stands for their total numbers. The concept-aware embeddings ${\mathbf{z}}_l^i \in\mathbb{R}^{d}$ are computed as:
\begin{equation}\label{eqn-1} 
\hat{\mathbf{z}}_l^i =\mathrm{Softmax}(\frac{\mathbf{q}_i\mathbf{k}_l^\top}{\sqrt{d_k}})\mathbf{v}_l,1\leq i\leq N,
\end{equation}
\begin{equation}\label{eqn-2} 
{\mathbf{z}}_l^i =LN(FFN(\hat{\mathbf{z}}_l^i )+\mathbf{q}_i),1\leq i\leq N,
\end{equation}
where $FFN(\cdot)$ denotes feed forward network and $LN(\cdot)$ denotes layer normalization. By performing this operation across all encoder layers, we obtain $L$ embeddings, which are then aggregated by a linear selector to produce multilayer embeddings $\mathbf{Z}^{i}$. In addition, for the $l^{th}$ layer, ${\mathbf{z}}_l^i $ also serves as input parallel to the image embeddings, as will be discussed below.

 \textbf{Concept prompt tuning (CPT).} Motivated by visual prompt tuning (VPT)~\cite{VPT}, we introduce concept prompt tuning. This design preserves the rich representations of pre-trained models while substantially addressing the performance degradation from excessive parameter tuning. Moreover, CPT supports concept-specific prompt design, thereby enabling self-attention mechanisms within each layer to progressively amplify target visual concepts, which significantly refines feature representations in a task-driven manner. 

 To be specific, let patch embeddings $\mathbf{P}_l=\left\{\mathbf{p}_l^{k_p}\in\mathbb{R}^d\mid {k_p}\in\mathbb{N},1\leq {k_p}\leq N_p\right\}$ of length $N_p$ and class token embedding $\mathbf{a}_l\in\mathbb{R}^d$ denoted as input to $l^{th}$ layer $\Phi_l$ of the original image encoder $\Phi(\cdot;\theta)$. We introduce concept-aware visual embeddings $\mathbf{Z}_l=\left\{\mathbf{z}_l^i\in\mathbb{R}^d\mid i\in\mathbb{N},1\leq i\leq N\right\}$, concatenated with $\mathbf{a}_l$ and $\mathbf{P}_l$, while keeping all backbone weights $\theta$ frozen. The entire process can be formulated as:
\begin{equation}\label{eqn-3} 
[\mathbf{a}_l,\_,\mathbf{P}_l]=\Phi_l\left([\mathbf{a}_{l-1},\mathbf{Z}_{l-1},\mathbf{P}_{l-1}]\right),l=1,2,\ldots,L.
\end{equation}
\subsection{Explainable Diagnose}
 For each concept, its candidate set $\mathcal{C}_i$ is formatted into structured text templates (e.g. ``\emph{This is a dermoscopic image, the} \{\emph{concept title}\} \emph{of the lesion is} \{$c_i^j$\}'') and encoded by a pre-trained text encoder into embeddings $T_{i}=[t_{i}^{1},...,t_{i}^{k_{i}}]\in\mathbb{R}^{k_{i}\times d}$, where $k_i$ is the length of $\mathcal{C}_i$. During the concept alignment phase, contrastive learning is applied between visual and textual concept features to maximize their semantic consistency through alignment loss formulated as:
\begin{equation}\label{eqn-4} 
\mathcal{L}_{con}=-\frac{1}{N}\sum_{i=1}^N\log{\frac{\exp (sim(\mathbf{Z}^i,t_i^*/\tau))}{\sum_{j=1}^{k_i}\exp (sim(\mathbf{Z}^i,t_i^j/\tau))}},
\end{equation}
where $t_i^*$ is the true label of concept $c_i$, $\tau$ is a learnable temperature parameter and $sim(\cdot,\cdot)$ represents cosine similarity. Afterwards, textual embeddings are fused into $D_i$ within the concept candidate set $\mathcal{C}_i$ according to their normalized similarity alignment scores for disease diagnosis.

 Emulating clinical decision-making workflows where experts synthesize multi-criteria assessments to derive conclusions and modified by~\cite{CBM}, gated aggregation module is employed to adaptively combine concept-aligned representations from all learned concepts through learnable weights $\alpha$ that quantify each concept’s diagnostic relevance:
\begin{equation}\label{eqn-5} 
\hat{y}=FC(\sum_{i=1}^N\alpha_i\cdot D_i),
\end{equation}
where $FC$ denotes full-connection layer to make the final decision and $D_i$ denotes the $i^{th}$ fused textual concept representation.

 During the training phase, the optimization process simultaneously minimizes a composite loss function comprising concept alignment loss and diagnostic cross-entropy loss to supervise both concept and disease classification:
\begin{equation}\label{eqn-6} 
\mathcal{L}=\lambda \mathcal{L}_{con}+(1-\lambda)\mathcal{L}_{CE}(\hat{y},y),
\end{equation}
where $\lambda \in [0,1]$ is a hyperparameter controlling the relative importance of concept and disease accuracy.
\section{Experiment}
\subsection{Experiment Setup}
\subsubsection{Datasets.} 
$\mathbf{PH^2}$~\cite{ph2} comprises 200 dermoscopic images, with annotations for five morphological features. By merging subcategories encompassing common and atypical nevi classes, the final dataset consists of 160 nevus and 40 melanoma cases. \textbf{Derm7pt}~\cite{derm7pt} consists of 1,011 dermoscopic image cases, with annotations based on the clinical 7-point checklist. Following Bie et al.~\cite{MICA}, we retain all seven clinical indicators with 575 nevi and 252 melanoma cases. \textbf{SkinCon}~\cite{skincon} contains 3,690 clinical images from Fitzpatrick 17k~\cite{F17k}. In this study, 22 high-frequency clinical features, each with over 50 annotation instances, are selected. The disease categories include non-neoplastic, benign, and malignant. All datasets are randomly divided into training, validation, and test sets in a 70\%:15\%:15\% ratio.
\begin{table}[t]
\centering
\footnotesize
\renewcommand{\arraystretch}{1.2}
\setlength{\tabcolsep}{1pt}
\caption{Quantitative comparison results in \textbf{disease prediction} in terms of area Under Curve (AUC), accuracy and F1-score. Results are in percentage(\%).}
\vspace{\captab}
\label{table:1}
\begin{tabular}{c|ccc|ccc|ccc}
\Xhline{1.2pt}
\multirow{2}{*}{Method} & \multicolumn{3}{c|}{$\mathbf{PH^2}$} & \multicolumn{3}{c|}{\textbf{Derm7pt}} &  \multicolumn{3}{c}{\textbf{SkinCon}}\\
                        & AUC        & ACC          & F1           & AUC        & ACC        & F1         & AUC        & ACC        & F1         \\ \hline
PCBM~\cite{PCBM}                   & $78.3_{{1.2}}$ & $89.3_{{1.9}}$   & $81.5_{{2.6}}$   & $73.0_{{2.2}}$ & $77.0_{{1.4}}$ & $71.0_{{1.2}}$ & $68.9_{{1.6}}$ & $71.0_{{1.1}}$ & $70.5_{{0.8}}$ \\
PCBM-h~\cite{PCBM}                  & $92.3_{{1.5}}$ & $90.7_{{1.9}}$   & $83.3_{{2.6}}$   & $83.3_{{1.1}}$ & $79.9_{{0.9}}$ & $74.5_{{1.4}}$ & $69.5_{{1.7}}$ & $72.3_{{1.4}}$ & $72.3_{{1.3}}$ \\
CBE~\cite{CBE}                     & $97.5_{{0.0}}$ & $96.0_{{0.0}}$   & $93.9_{{0.0}}$   & $76.6_{{0.4}}$ & $83.8_{{0.3}}$ & $78.1_{{0.4}}$ & $72.8_{{1.2}}$ & $73.8_{{1.1}}$ & $73.6_{{1.3}}$ \\
Explicd~\cite{Explicd}                 & $95.4_{{2.4}}$ & $94.4_{{2.3}}$   & $92.8_{{3.6}}$   & $87.5_{{3.2}}$ & $81.0_{{3.2}}$ & $80.5_{{3.5}}$ & $74.0_{{0.9}}$ & $73.1_{{0.3}}$ & $72.0_{{0.7}}$ \\
MICA~\cite{MICA}                    & $98.2_{{1.4}}$ & $98.7_{{1.9}}$   & $95.3_{{1.2}}$   & $85.6_{{1.1}}$ & $83.9_{{1.0}}$ & $79.4_{{1.3}}$ & $75.9_{{1.1}}$ & $75.6_{{1.1}}$ & $75.4_{{1.2}}$ \\ \hline
\textbf{Our work}       & \bm{$98.3_{{1.6}}$} & \bm{$98.9_{{2.2}}$} & \bm{$98.8_{{2.3}}$} & \bm{$92.1_{{1.2}}$} & \bm{$86.0_{{0.5}}$} & \bm{$85.8_{{0.9}}$} & \bm{$77.5_{{0.6}}$} & \bm{$76.3_{{0.8}}$} & \bm{$75.7_{{0.8}}$}      \\
\Xhline{1.2pt}
\end{tabular}
\vspace{\tabbottom}
\end{table}
\subsubsection{Implementation Details.} Our framework initializes both image and text encoders using BiomedCLIP~\cite{biomedclip} pre-trained weights and settings. Optimization is performed using Adam with a learning rate $\eta$=1e-5. The loss weighting coefficient $\lambda$ controlling concept-task balance is set to 0.5 via cross-validation. All experiments are implemented in PyTorch and executed on NVIDIA GeForce RTX 3090 GPUs, with results averaged over three random seeds.
\subsection{Experiment Result}
\subsubsection{Comparison with existing methods.} 
To evaluate the efficacy of our approach, we compare it with existing concept-based methods on the aforementioned datasets. Methods for comparison include: PCBM~\cite{PCBM}, CBE~\cite{CBE}, Explicd~\cite{Explicd}, and MICA~\cite{MICA}. Table \ref{table:1} and \ref{table:2} summarize results for concept alignment and disease classification, respectively. Our framework achieves state-of-the-art performance across most metrics, especially on Derm7pt, where CoPA's disease prediction accuracy exceeds the second-best by 2.1\%, and on PH$^2$ with a concept prediction accuracy 2.6\% higher than the second-best approach. 
\subsubsection{Ablation study.} We conduct various ablation studies on PH$^2$ and Derm7pt to investigate the influence of different modules and settings. Table \ref{table:3} quantifies the contribution of individual components to overall performance. Specifically, our ablation analysis show that all designed components contribute positively, including 1) MultiLayer Aggregation strategy (MLA), which hierarchically aggregates features across encoder layers to capture multiscale features; 2) Concept Prompt Tuning (CPT), designed to highlight the concept-specific feature; and 3) Frozen Vision Backbone (FVB) that address knowledge forgetting caused by parameter-efficient fine-tuning. Table \ref{table:3} shows that our approach, including all three components, achieves optimal performance.
\begin{table}[thbp]
\centering
\footnotesize
\renewcommand{\arraystretch}{1.3}
\setlength{\tabcolsep}{6pt}
\caption{Quantitative comparison results in \textbf{concept prediction} in terms of AUC, accuracy and F1-score. Results are in percentage(\%).}
\vspace{\captab}
\label{table:2}
\begin{tabular}{c|ccc|ccc|ccc}
\Xhline{1.2pt}
\multirow{2}{*}{Method} & \multicolumn{3}{c|}{PH$^2$}                    & \multicolumn{3}{c|}{Derm7pt}                  & \multicolumn{3}{c}{SkinCon}                   \\
                        & AUC         & ACC           & F1            & AUC           & ACC           & F1            & AUC           & ACC           & F1            \\ \hline
CBE~\cite{CBE}                     & 81.3        & 71.6          & 70.0          & 72.2          & 74.1          & 71.0          & 79.3          & 89.0          & 62.1          \\
Explicd~\cite{Explicd}                 & 88.8        & 79.6          & 76.4          & 85.7          & 73.0          & 71.9          & 76.0          & 93.1          & 65.5          \\
MICA~\cite{MICA}                    & 83.6        & 75.2          & 68.4          & 78.6          & 76.0          & 72.4          & \textbf{82.6} & 91.7          & 63.8          \\ \hline
\textbf{Our work}                & \textbf{89.0} & \textbf{82.2} & \textbf{80.6} & \textbf{87.0} & \textbf{77.1} & \textbf{76.6} & 81.7          & \textbf{93.6} & \textbf{70.4} \\ \hline
\Xhline{1.2pt}
\end{tabular}
\vspace{\tabbottom}
\end{table}
\begin{table}[]
\centering
\footnotesize
\renewcommand{\arraystretch}{1.3}
\setlength{\tabcolsep}{6pt}
\caption{Ablation study of CoPA on PH$^2$ and Derm7pt. MLA, CPT and FVB represent Multi-Layer Aggregation, Concept Prompt Tuning and Freezing Vision Backbone.}
\vspace{\captab}
\label{table:3}
\begin{tabular}{ccc|cccccccc}
\Xhline{1.2pt}
\multirow{3}{*}{MLA} & \multirow{3}{*}{CPT} & \multirow{3}{*}{FVB} & \multicolumn{4}{c}{PH$^2$}                                 & \multicolumn{4}{c}{Derm7pt}                             \\ \cline{4-11} 
                     &                      &                      & \multicolumn{2}{c}{Label} & \multicolumn{2}{c}{Concept} & \multicolumn{2}{c}{Label} & \multicolumn{2}{c}{Concept} \\
                     &                      &                      & ACC         & F1          & ACC          & F1           & ACC         & F1          & ACC          & F1           \\ \hline
\XSolidBrush         & \XSolidBrush         & \XSolidBrush         & 93.3        & 93.9        & 79.6         & 76.4         & 81.7        & 81.2        & 73.0         & 71.9         \\
\CheckmarkBold       & \XSolidBrush         & \XSolidBrush         & 96.7        & 96.5        & 80.9         & 78.1         & 85.0        & 84.2        & 73.1         & 72.2         \\
\XSolidBrush         & \CheckmarkBold       & \XSolidBrush         & 96.7        & 95.2        & 80.0         & 76.6         & 84.5        & 85.0        & 73.6         & 73.1         \\
\XSolidBrush         & \CheckmarkBold       & \CheckmarkBold       & 96.7        & 96.7        & 80.5         & 81.7         & 85.0        & 85.2        & 76.0         & 75.2         \\
\CheckmarkBold       & \CheckmarkBold       & \XSolidBrush         & 97.8        & 96.4        & 81.1         & 79.1         & 85.2        & 85.5        & 74.0         & 73.2         \\
\CheckmarkBold       & \CheckmarkBold       & \CheckmarkBold       & \textbf{98.9}   & \textbf{98.8}   & \textbf{82.2}    & \textbf{80.6}    & \textbf{86.0}   & \textbf{85.8}   & \textbf{77.1}    & \textbf{76.6}    \\ 
\Xhline{1.2pt}
\end{tabular}
\vspace{\tabbottom}
\end{table}

\subsection{Interpretability Analysis}
Inspired by previous work~\cite{guidotti2018survey,hsiao2021roadmap,rigotti2021attention}, we analyze the interpretability of our model from the following three aspects: 
\textit{faithfulness}, \textit{understandability}, and \textit{plausibility}. 
\subsubsection{Faithfulness.} 
\begin{figure}[thbp]
\centering
\begin{minipage}{.4\textwidth}
\centering\includegraphics[width=\textwidth]{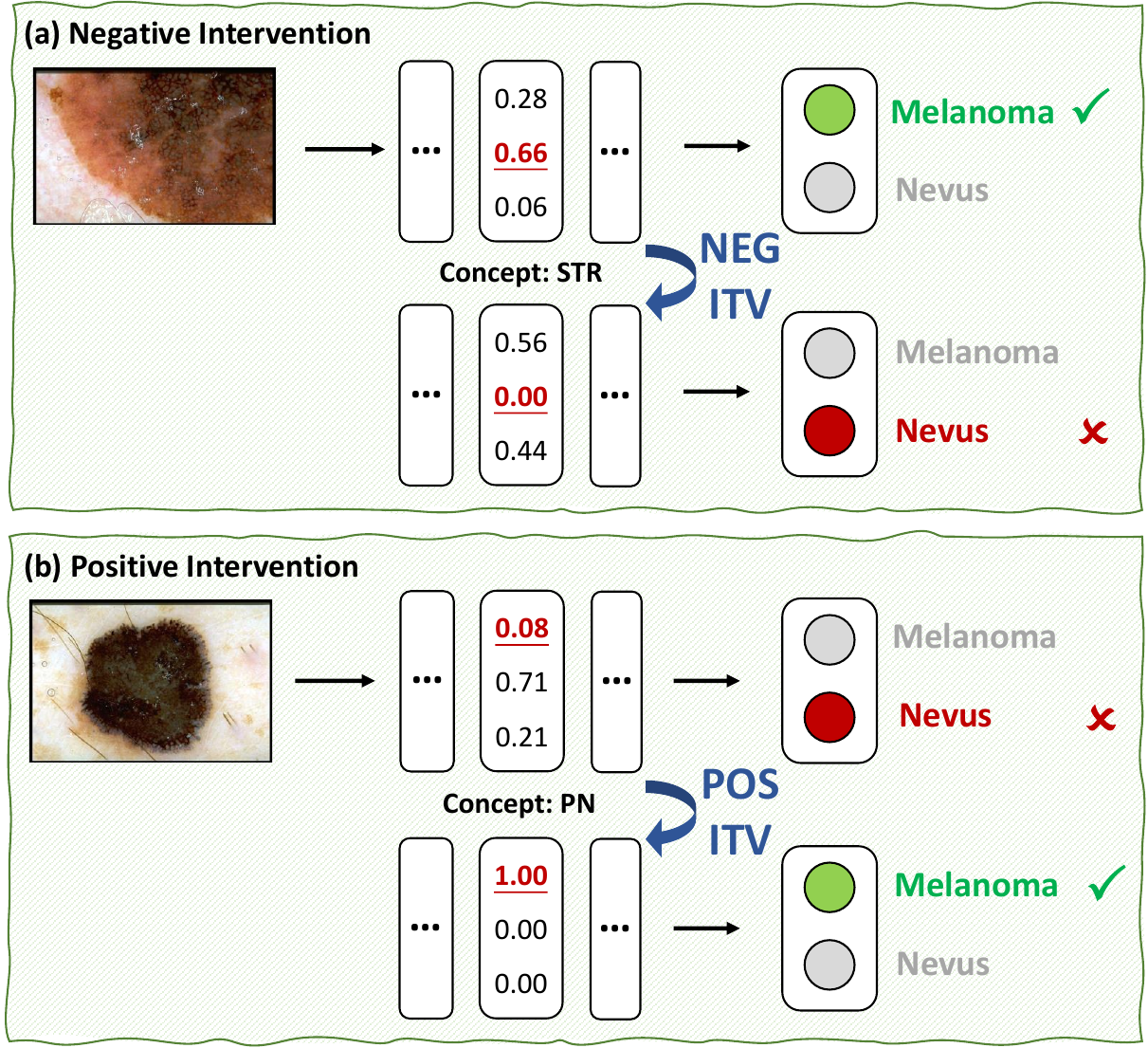}
\vspace{-0.7cm}
\caption{Intervention Examples. ITV: Intervention}
\label{fig:intervention}
\end{minipage}
\hfill
\begin{minipage}{.55\linewidth}
\begin{table}[H]
\vspace{-0.3cm}
\centering
\footnotesize
\renewcommand{\arraystretch}{1.6}
\setlength{\tabcolsep}{3pt}
\begin{tabular}{cccc}
\Xhline{1.2pt}
                          & \textbf{ITV nums} & \textbf{ACC} & \textbf{Improve}                     \\ \hline
ITV-Free                  & 0        & 86      & -                           \\ \hline
                          & 1        & 83.6    & {\color[HTML]{3531FF} -2.4} \\
\multirow{-2}{*}{Neg ITV} & 2        & 82.8    & {\color[HTML]{3531FF} -3.2} \\ \hline
                          & 1        & 86.5    & {\color[HTML]{CB0000} 0.5}  \\
\multirow{-2}{*}{Pos ITV} & 2        & 87.1    & {\color[HTML]{CB0000} 1.1}  \\ \hline
\Xhline{1.2pt}
\end{tabular}
\vspace{0.8cm}
\caption{Accuracy changes of test-time concept intervention on Derm7pt.}
\label{table:4}
\end{table}
\end{minipage}
\vspace{\tabbottom}
\end{figure}

\textit{Faithfulness} refers to the extent to which explanations accurately reflect the internal decision-making process of the model~\cite{guidotti2018survey,rigotti2021attention}. In this study, we evaluate model's faithfulness through manual intervention during inference on Derm7pt. Specifically, as shown in Fig. \ref{fig:intervention}, for \underline{positive intervention}, we set 1-2 incorrectly predicted concepts' ground-truth confidence to 1, while for \underline{negative intervention}, we adjust 1-2 correctly predicted concepts' ground-truth confidence to 0, to observe the outcome changes of the disease prediction. Notably, other confidence of adjusted concepts is recalculated by softmax function to ensure the probability sum equals 1. As shown in Table \ref{table:4}, positive interventions resulted in accuracy increase of 0.5\% (single-concept) and 1.1\% (two-concept), whereas negative interventions led to accuracy reductions by 2.4\% and 3.2\%, respectively, showing model's heavy reliance on concept predictions and affirming the faithfulness of the explanations.
\subsubsection{Understandability \& Plausibility.} 
\begin{figure}[thbp]
\centering
\subfigure[]{
    \begin{minipage}{.44\textwidth}
        \centering
        \includegraphics[width=\linewidth]{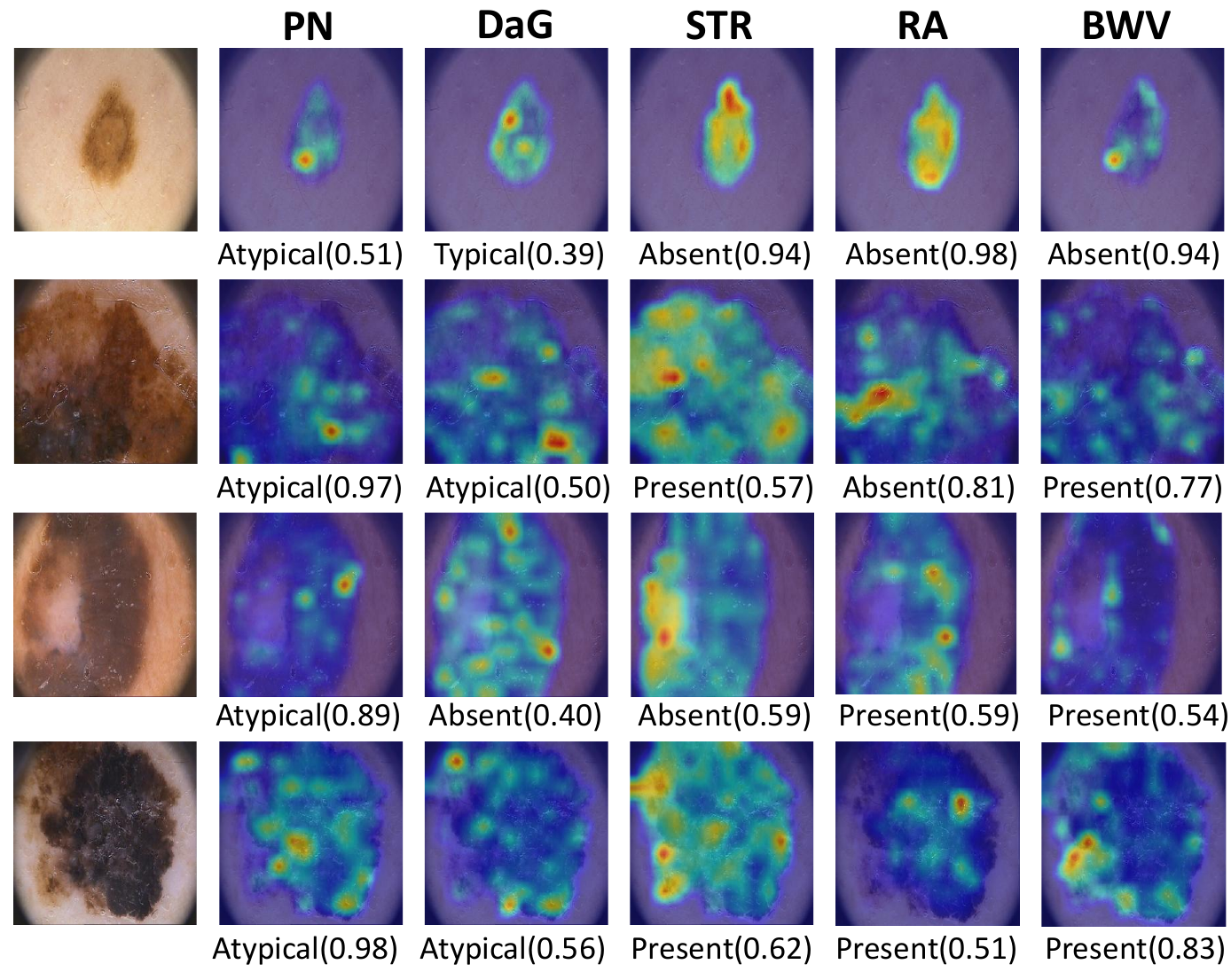}
    \end{minipage}
}
\hfill
\subfigure[]{
    \begin{minipage}{.50\linewidth}
        \centering
        \includegraphics[width=\linewidth]{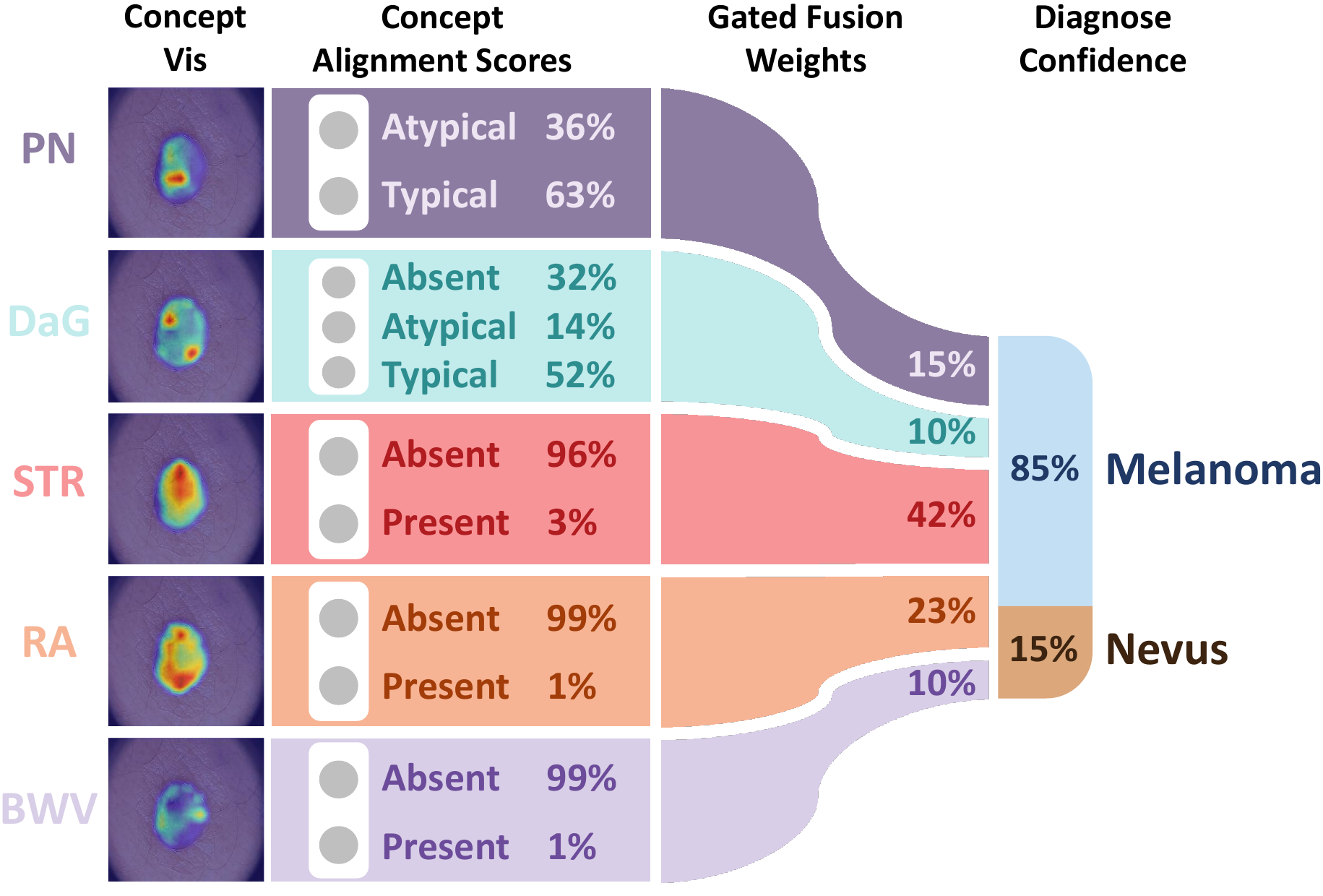}
    \end{minipage}
}
\caption{Illustration of understandability and plausibility, where PN, DaG, STR, RA, BWV stand for ``Pigment Network'', ``Dots and Globules'', ``Streaks'', ``Regression Area'', ``Blue-Whitish Veil'', respectively.   
(a) Heatmap visualization of concern areas for each concept. (b) The entire process of the prediction, including concept visualization, concept alignment scores, gated fusion mechanism weights, and diagnose confidence.}
\label{fig:understandability}
\end{figure}

$\mathit{Understandability}$ refers that the explanation's context should be readily comprehensible to users, eliminating the necessity for technical expertise~\cite{jin2023guidelines}, while $\mathit{plausibility}$ refers to the extent to which explanations align with domain-specific human reasoning and appear credible~\cite{carvalho2019machine}. Fig. \ref{fig:understandability} shows the examples of explanations in detail. In Fig. \ref{fig:understandability}(a), we visualize concept-associated regions and their prediction confidence scores, providing users with a foundation to assess the acceptability of concept alignment. Fig. \ref{fig:understandability}(b) illustrates the process of concept prediction and disease diagnosis for a data sample, including visual concept heatmaps, concept confidence scores, gated network weights, and diagnostic confidence. This workflow provides users with a transparent and traceable insight into the entire prediction process, ensuring the interpretability of diagnostic decisions.



\section{Conclusion}
In this paper, we propose CoPA, a multilayer concept prompting and aggregation framework for interpretable disease diagnosis. 
Within the Concept-aware Embedding Generator (CEG) of this framework, concept anchors is employed to query multi-scale visual features, generating densely concentrated concept representations that are hierarchically aggregated. 
Furthermore, to preserve the vast knowledge from the pre-trained vision-language model while enabling discriminative fine-tuning, we introduce Concept Prompt Tuning (CPT), which utilizes concept-aware representations as task-oriented visual prompts, guiding the model to focus on concept-relevant features. 
Experiments on three datasets demonstrate the exceptional performance and interpretability of our method.
\begin{credits}
\subsubsection{Acknowledgments.} This work was supported in part by the Natural Science Foundation of China under Grant 62472184, and in part by the Fundamental Research Funds for the Central Universities.
\subsubsection{Disclosure of Interests.}The authors have no competing interests to declare that are relevant to the content of this article.
\end{credits}
\bibliographystyle{splncs04}
\bibliography{ref.bib}
\end{document}